\pdfoutput=1
\documentclass[11pt]{article}
\usepackage[preprint]{acl}
\usepackage{times}
\usepackage{latexsym}
\usepackage[T1]{fontenc}
\usepackage[utf8]{inputenc}
\usepackage{microtype}
\usepackage{inconsolata}
\usepackage{graphicx}
\usepackage{amsmath}
\usepackage{booktabs}
\usepackage{makecell}

\title{Equity with Efficiency: An Empirical Study of Tokenizers for \\ Multilingual Large Language Models}

\author{
    \textbf{Kieron Seven Jun Wei Lee\textsuperscript{1}}
    \textbf{Muhammad Reza Qorib\textsuperscript{2}}
    \\
    \textbf{Andrew Ivan Soegeng\textsuperscript{1,3}}
    \textbf{Hwee Tou Ng\textsuperscript{1}}
    \\
    \textsuperscript{1}National University of Singapore
    \textsuperscript{2}Carnegie Mellon University
    \textsuperscript{3}SAP
    \\
    \texttt{e0968891@u.nus.edu},
    \texttt{mrqorib@cmu.edu},
    \\
    \texttt{andrew.soegeng@u.nus.edu}, \texttt{dcsnght@nus.edu.sg}
}

\begin{document}
\maketitle
\begin{abstract}
Multilingual large language models (LLMs) depend on subword tokenization to bridge discrete text and continuous neural representation. State-of-the-art multilingual LLMs often use Byte-level Byte-Pair Encoding (BPE) tokenizers that structurally favor high-resource languages and Latin scripts. For speakers of underrepresented languages, particularly those across Southeast Asia, this bias inflates inference costs and widens cross-lingual capability gaps. We present the first systematic comparison of equitable tokenizers on a unified benchmark spanning 11 Southeast Asian languages. Beyond tokenizer-level analysis of compression efficiency and cross-lingual equity, we assess downstream task performance through controlled 1.5B-parameter language model training using the same training data. Our results show that Parity-aware BPE lies on the Pareto frontier of the efficiency-equity trade-off, achieving strong compression parity at competitive cost. Morphology-Driven Byte Encoding delivers the best semantic reasoning performance through morphologically richer representations, albeit at a higher computational expense. Byte Latent Transformer underperforms on downstream tasks, possibly because its architectural assumptions misalign with the constraints of limited low-resource training data. Together, our findings demonstrate that cross-lingual fairness and tokenization efficiency are not fundamentally at odds, and offer practical guidance for designing equitable multilingual models.\footnote{Source code will be publicly released upon paper publication.}
\end{abstract}

\section{Introduction}
Multilingual large language models (LLMs) are central to cross-lingual information access, yet their performance remains deeply uneven across languages and scripts. A key driver of this disparity is tokenization: how raw text is segmented into subword units shapes model capacity, sequence length, and effective context window across languages \citep{petrov2023languagemodeltokenizersintroduce} .

Byte-level Byte-Pair Encoding (BPE) \citep{sennrich2016neuralmachinetranslationrare} is a widely used tokenization strategy in state-of-the-art LLMs, including the GPT \citep{OpenAI} and Llama \citep{touvron2023llama2openfoundation} families, due to its simplicity and compression efficiency. Byte-level BPE encodes characters as UTF-8 bytes \citep{unicode2011} and iteratively learns byte-pair merges based on global co-occurrence frequency. This procedure introduces a structural bias as one Latin character is encoded as a single byte, while one non-Latin character requires two or more bytes. Combined with English-centric pretraining corpora, BPE's merge operations disproportionately favor Latin scripts and high-resource languages \citep{arnett2024bitproblemmeasurementdisparities}. 

The practical consequences of such bias are significant. \citet{petrov2023languagemodeltokenizersintroduce} demonstrated that GPT-4's Byte-level BPE tokenizer produces sequence length disparities of up to 15×, with Chinese requiring 1.9× more tokens than English, Vietnamese 2.5×, and Burmese 11.7×. For speakers of low-resource non-Latin languages such as Khmer and Lao, these disparities translate directly into higher inference costs, degraded long-context reasoning, and diminished downstream task accuracy \citep{tamang2024evaluatingtokenizerperformancelarge}.

Several tokenizers have been proposed to address these inequities. Parity-aware Byte-Pair Encoding rebalances merge frequencies across scripts \citep{foroutan2025parityaware}. Morphology-Driven Byte Encoding (MYTE) grounds segmentation in morphological structure \citep{Limisiewicz_2024}. Byte Latent Transformer (BLT) sidesteps a fixed vocabulary by operating directly over dynamic byte patches \citep{pagnoni2024bytelatenttransformerpatches}. Each work evaluates its approach against BPE baselines, reporting improvements in equity and multilingual capability. However, these methods have never been compared against each other under uniform experimental conditions.

In this paper, we present a benchmarking study to address this gap with the first systematic analysis of equitable tokenizers. We compare them across eleven Southeast Asian (SEA) languages: English, Burmese, Chinese, Indonesian, Khmer, Lao, Malay, Tagalog, Tamil, Thai, and Vietnamese. Using Byte-level BPE as a baseline, and controlling for training data, vocabulary size, and computational budget, we evaluate intrinsic tokenizer metrics and examine downstream LLM performance by training 1.5B-parameter decoder-only language models from scratch. Our study provides a direct empirical comparison of equitable tokenization methods, offering actionable insights for NLP practitioners to build fairer multilingual LLMs.

\section{Related Work}

\subsection{Subword Tokenization}

Subword tokenization has become the standard preprocessing step in multilingual LLMs, to uniformly segment text in any language into tokens. However, when trained on heterogeneous multilingual corpora, these approaches allocate vocabulary capacity toward languages with high resource or written in Latin scripts, embedding structural bias and inequity into the vocabulary.

The downstream consequences are well-documented. \citet{bostrom2020bytepairencodingsuboptimal} showed that BPE tokens frequently diverge from linguistically motivated morpheme boundaries. More recently, \citet{selvamurugan2025from} quantified cross-lingual tokenization inequity through normalized sequence length and subword fertility, demonstrating that the gap is most pronounced for underrepresented scripts. These findings motivate moving beyond global frequency optimization as the main design criterion for multilingual tokenizers.

\subsection{Parity-aware Byte-Pair Encoding}
Parity-aware BPE (PA BPE; \citealp{foroutan2025parityaware}) modifies Byte-level BPE by optimizing the worst-case compression rate across languages. Each merge iteration selects the pair that most improves the worst-performing language, trading marginal global efficiency for tokenization equity.

The approach requires minimal implementation changes to existing BPE pipelines. On a 30-language unbalanced dataset, it achieves a lower Gini coefficient of 0.011 versus 0.064 for Byte-level BPE, while remaining competitive on compression and outperforming or matching Byte-level BPE baselines across 13 multilingual benchmarks.

\subsection{Morphology-Driven Byte Encoding}

MYTE \citep{Limisiewicz_2024} replaces UTF-8's character-based convention with morpheme-based byte codes, as morphemes exhibit more consistent sequence lengths than characters across languages. It learns a per-language morpheme inventory to achieve balanced morphological coverage via Morfessor 2.0 \citep{smit-etal-2014-morfessor}, and assigns shorter byte sequences to linguistically meaningful units.

MYTE produces shorter encoding compared to UTF-8 for all 99 languages tested, with gains ranging from 1\% for Vietnamese and Chinese to nearly 70\% for Burmese. Its worst-case tokenizer parity relative to English is 1.7, versus 3.5 for UTF-8. MyT5, a MYTE-encoded variant of ByT5 \citep{xue2022byt5tokenfreefuturepretrained}, demonstrated reduced cross-language perplexity disparity compared to its byte-level counterpart. It achieves 75.3 F1 on XTREME-UP \citep{Ruder_2023} question answering versus 73.2 for ByT5.

\subsection{Byte Latent Transformer}
BLT \citep{pagnoni2024bytelatenttransformerpatches} eliminates explicit tokenization entirely and comprises three modules: a lightweight local encoder producing patches, a large latent transformer processing them, and a lightweight local decoder reconstructing bytes. An entropy model drives patch segmentation, allocating computation proportional to data complexity.

BLT enables a 50\% reduction in inference FLOPs relative to Llama 3’s original tokenizer without sacrificing downstream task performance. \citep{grattafiori2024llama3herdmodels}.
By avoiding a static vocabulary from tokenization, BLT sidesteps multilingual inequity that arises when high-resource language tokens dominate and outperforms Llama 3 by 2 BLEU points \citep{papineni-etal-2002-bleu} on translation into English.

\section{Methods}

We compare the three tokenizer families discussed above to a baseline Byte-level BPE tokenizer. We train all tokenizers on the same dataset to evaluate their efficiency and cross-lingual equity. We then train language models from scratch using these tokenizers and evaluate their downstream task performance. For fairness and reproducibility, data sizes are reported in number of sentences and bytes, rather than tokens.

\subsection{Training Data}

For tokenizer training, we sample a total of 1 million sentences (3.5GB) across eleven SEA languages from multilingual C4 (mC4) \citep{xue-etal-2021-mt5}. Sampling is performed randomly without replacement following the language proportions in mC4 to approximate realistic multilingual data distribution. The resulting per-language sentence counts are detailed in Appendix~\ref{app:tok-training-data}. 

For language model training, we adopt the same training dataset as \citet{foroutan2025parityaware} and sample 100 million sentences (203 GB) from FineWeb2 \citep{penedo2025fineweb2pipelinescale}. This dataset size is comparable to what \citet{foroutan2025parityaware} and  \citet{Limisiewicz_2024} used to train their language models. FineWeb2 is a multilingual web corpus with quality filtering already applied, and we did not apply further preprocessing before training. Language proportions are controlled using temperature sampling with $\tau=1.21$ to boost the representation of low-resource languages \citep{foroutan2025parityaware}. The details are provided in Appendix~\ref{app:lm-training-data}. 

Vocabulary sizes are controlled where possible to enable a fair comparison of the four tokenizers. MYTE was designed to have 4,096 morphemes per language to avoid over-segmentation. Thus, we train tokenizers at three scales: 4,096, 8,192, and 12,288 tokens per language, across all eleven SEA languages. For MYTE, this translates to total morpheme inventories of $45k$, $90k$, and $135k$ morphemes. The vocabulary sizes of Byte-level BPE and Parity-aware BPE are matched to MYTE's total morpheme counts at each scale. 

BLT's patch-based representation is not directly comparable since it does not learn a fixed vocabulary. Following the approach of \citet{pagnoni2024bytelatenttransformerpatches}, we configure BLT's entropy model to yield average patch sizes of 4.5, 6, and 8 bytes per patch.

We use tokenizers with vocabulary size of $90k$ to train language models, placing them close to the $100k$--$128k$ vocabulary size of most LLM tokenizers \citep{wegmann2025tokenizationsensitivelanguagevariation}. For BLT, we adopt the entropy model with an average patch size of 4.5 bytes, following the setup of \citet{pagnoni2024bytelatenttransformerpatches}. Note that BLT is not a tokenizer in the traditional sense, but is referred to as one here for ease of comparison.

\subsection{Implementation Details}

Training of tokenizers and tokenization of language model training data for MYTE and BPE-based algorithms were performed on a single AMD EPYC 9554P CPU (128 threads). For BLT, the entropy-based tokenizer was trained on 4× NVIDIA H100 GPUs, and the language model training dataset was tokenized on 8× NVIDIA H200 GPUs. Statistics of the language model training dataset after tokenization are reported in Table~\ref{tab:tokenization-stats}.

{\setlength{\tabcolsep}{5pt}
\begin{table}[ht!]
    \centering
    \begin{tabular}{lccc}
        \toprule
        Tokenizer & Duration & \# Tokens & File size \\
        (size) & (hour) & (billion) & (GB) \\
        \midrule
        BLT (4.5)    & 33 &  42 & 204 \\
        MYTE (90k)   & 50 & 269 & 538 \\
        PA BPE (90k) &  3 &  82 & 329 \\
        BPE (90k)    &  3 &  72 & 288 \\
        \bottomrule
    \end{tabular}
    \caption{Statistics of language model training dataset after tokenization by the four tokenizers. Legend: size = patch size for BLT, morpheme inventory size for MYTE, vocabulary size for all other models; File Size = Size of dataset files after tokenization; PA BPE = Parity-aware BPE; BPE = Byte-level BPE.}
    \label{tab:tokenization-stats}
\end{table}
}

Language model training is carried out on 4--8× NVIDIA H100/H200 GPUs. To enable a fair comparison of computational cost, training durations are converted to an 8× NVIDIA H200 equivalent, as reported in Table~\ref{tab:lm-training-durations}. MYTE incurs the highest training cost at 300 normalized hours due to its substantially larger token count (269B tokens), while Byte-level BPE is the most efficient at 68 hours (72B tokens). Additionally, we trained and compared all language models at an equal token count of 38B tokens as measured by their respective tokenizers. These experiments yielded the same conclusions as the models trained on the same dataset, so we omit them for the sake of brevity.

\begin{table}[ht!]
    \centering
    \begin{tabular}{lccc}
        \toprule
        Model & Duration & \# Tokens  \\
        (size) & (hour) & (billion)\\
        \midrule
        BLT (4.5)    & 160 &  42 \\
        MYTE (90k)   & 300 & 269 \\
        PA BPE (90k) &  87 &  82 \\
        BPE (90k)    &  68 &  72 \\
        \bottomrule
    \end{tabular}
    \caption{Statistics of language model training.}
    \label{tab:lm-training-durations}
\end{table}

\subsection{Evaluation Metrics}
\subsubsection{Intrinsic Metrics}

Quantifying tokenizer efficiency and cross-lingual fairness requires metrics that are agnostic to both language and model architecture. We identify three such metrics from recent literature and provide brief descriptions below. Detailed definitions and formulae can be found in Appendix~\ref{intrin-tok}.

\textbf{Tokenizer parity} measures the ratio of the number of tokens per sentence in a given language relative to English \citep{petrov2023languagemodeltokenizersintroduce}. A \emph{tokenizer parity close to 1} indicates that the tokenizer imposes roughly equal computational cost across the given language and English.

\textbf{Gini coefficient} adapts the income inequality measure to the domain of tokenization fairness \citep{foroutan2025parityaware}. It quantifies the distribution of per-language tokenization costs, with values ranging from 0 (perfect equality) to 1 (maximal inequality). A \emph{lower Gini coefficient} reflects a more equitable tokenizer. 

\textbf{Compression rate} measures how efficiently a tokenizer compresses text \citep{foroutan2025parityaware}. A \emph{higher compression rate} indicates that the tokenizer is more efficient and produces fewer tokens for the same amount of text.

\subsubsection{Extrinsic Metrics}

We evaluate trained language models on English and multilingual classification benchmarks using Language Model Evaluation Harness \citep{eval-harness} with zero-shot prompting. Details of the benchmarks can be found in Appendix~\ref{ext-met}. For machine translation, we evaluate fine-tuned models with five-shot prompts drawn from the training dataset of the multi-way parallel FLORES+ corpus \citep{costajussa2024nllb}, following the setup of \citet{Limisiewicz_2024}.

To assess English language understanding, models are evaluated on three English classification benchmarks used by \citet{pagnoni2024bytelatenttransformerpatches}: PIQA \citep{bisk2019piqareasoningphysicalcommonsense}, HellaSwag \citep{zellers2019hellaswagmachinereallyfinish}, and Arc-C \citep{clark2018thinksolvedquestionanswering}. These benchmarks test commonsense reasoning, sentence completion, and science question answering respectively. 

Cross-lingual generalization is assessed through three multilingual classification benchmarks used by \citet{foroutan2025parityaware}. XNLI \citep{conneau2018xnlievaluatingcrosslingualsentence} evaluates natural language inference across multiple languages, XCOPA \citep{ponti2020xcopamultilingualdatasetcausal} tests causal commonsense reasoning in a multilingual setting, and XStoryCloze \citep{lin2022fewshotlearningmultilinguallanguage} assesses story completion across languages.

Machine translation is assessed after fine-tuning via continual pre-training to ensure that results reflect the task-adapted performance of the models. Fine-tuning details and the resulting per-language sentence counts of the fine-tuning dataset are provided in Appendix~\ref{app:ft-training-data}. BLEU \citep{papineni-etal-2002-bleu} and chrF \citep{popovic-2015-chrf} scores are computed in both EN $\rightarrow$ XX and XX $\rightarrow$ EN directions for all ten non-English SEA languages. These two metrics range from 0 to 100 and higher scores indicate better translation quality.

\section{Experiments}
\subsection{Intrinsic Evaluation}
We evaluate the trained tokenizers on FLORES+ devtest set, which consists of 1,012 aligned sentences across all eleven SEA languages.
For Parity-aware BPE, we train the base variant with the FLORES+ training dataset as the development corpus following the setup of \citet{foroutan2025parityaware}. It is the only tokenizer among those evaluated that requires parallel data during training.

\subsection{Extrinsic Evaluation}
\subsubsection{Language Model}
The base architecture of our language model is OLMo-2-1B \citep{olmo20252olmo2furious}, a decoder-only transformer comprising 16 layers, a hidden dimension of 2,048, and approximately 1.5 billion parameters. We train all models from scratch to ensure that differences in downstream task performance are largely attributed to tokenizer choice.

\subsubsection{Training Configuration}
Models are trained using the AdamW \citep{loshchilov2018decoupled} optimizer with a peak learning rate of $4.0 \times 10^{-4}$, weight decay of 0.1, $\beta_1 = 0.9$, $\beta_2 = 0.95$, and gradient clipping of 1.0. The learning rate follows a Warmup-Stable-Decay (WSD) schedule \citep{hu2024minicpmunveilingpotentialsmall}: a linear warmup over the first 1\% of training tokens, a stable phase over the next 89\%, and a linear decay over the final 10\%. The global batch size is 512 sequences with a maximum sequence length of 4,096 tokens, corresponding to approximately 2 million tokens per training step.

\subsubsection{Statistical Significance}
Extrinsic metrics are assessed for statistical significance using paired bootstrap resampling with 1,000 iterations at $p < 0.05$ \citep{koehn_2004_statistical}. This ensures that reported performance differences between models reflect meaningful systematic effects rather than sampling variation across examples.

\section{Results}
\subsection{Intrinsic Evaluation}

\begin{table}[ht!]
    \centering
    \begin{tabular}{lccc}
        \toprule
        Tokenizer & CR & Gini & TP \\
        (size) & & &  \\
        \midrule
        BLT (4.5) & 0.0127 & 0.212 & 2.87 \\
        BLT (6) & 0.0145 & 0.219 & 2.96 \\
        BLT (8) & 0.0161 & 0.227 & 3.06 \\
        \midrule
        \makecell[l]{MYTE (45k)} & 0.0085 & 0.085 & \underline{1.23} \\
        \makecell[l]{MYTE (90k)} & 0.0089 & 0.086 & \underline{1.23} \\
        \makecell[l]{MYTE (135k)} & 0.0089 & 0.095 & 1.26 \\
        \midrule
        \makecell[l]{PA BPE (45k)} & 0.0250 & \textbf{0.021} & \textbf{1.15} \\
        \makecell[l]{PA BPE (90k)} & 0.0272 & \underline{0.028} & 1.24 \\
        \makecell[l]{PA BPE (135k)} & 0.0280 & 0.029 & 1.25 \\
        \midrule
        \makecell[l]{BPE (45k)} & 0.0257 & 0.243 & 1.93 \\
        \makecell[l]{BPE (90k)} & \underline{0.0293} & 0.220 & 1.73 \\
        \makecell[l]{BPE (135k)} & \textbf{0.0314} & 0.203 & 1.61 \\
        \bottomrule
    \end{tabular}
    \caption{Intrinsic evaluation of tokenizers on identical training data. Compression rate and tokenizer parity values are macro-averaged across languages. The best result is in \textbf{bold} and the second-best is \underline{underlined}. Legend: CR = Compression Rate, TP = Tokenizer Parity.}
    \label{tab:tokenizer-comparison}
\end{table}

\begin{figure*}[ht!]
    \centering
    \includegraphics[width=0.95\textwidth]{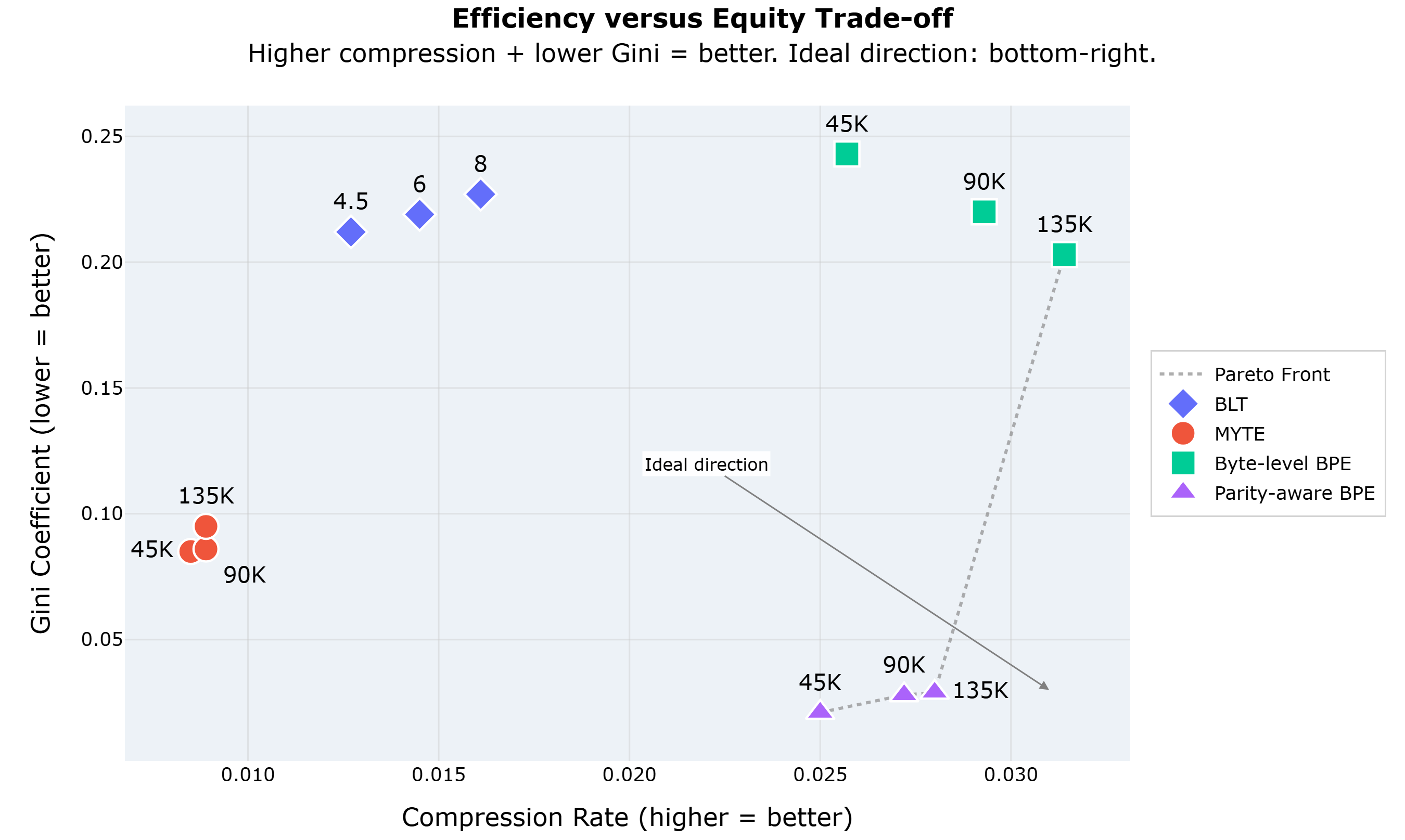}
    \caption{Efficiency-equity Pareto front of the evaluated tokenizers. Values beside markers indicate the patch size for BLT, morpheme inventory size for MYTE, and vocabulary size for BPE and PA BPE.}
    \label{fig:pareto}
\end{figure*}

Table~\ref{tab:tokenizer-comparison} reveals that Parity-aware BPE achieves the lowest Gini coefficient across all vocabulary sizes. This equity gain does not come at the cost of tokenizer efficiency, as Parity-aware BPE achieves competitive compression rates relative to Byte-level BPE.

MYTE has a relatively low Gini coefficient and tokenizer parity, but its compression rate is the worst. This indicates that morpheme-based segmentation produces longer token sequences. The trade-off is consistent with the inflated token counts observed during language model training.

BLT exhibits poor equity across all vocabulary sizes despite operating at the byte level without a fixed vocabulary. Its tokenizer parity is the highest among all approaches, suggesting that entropy-driven patch segmentation provides no built-in mechanism to correct for corpus imbalances.

Figure~\ref{fig:pareto} situates each tokenizer family within a two-dimensional efficiency-equity space, where the ideal direction is toward a higher compression rate and lower Gini coefficient (bottom-right of the plot). Parity-aware BPE lies on the Pareto front of the efficiency-equity space across all vocabulary sizes, highlighting that cross-lingual fairness and compression efficiency are not at odds. Byte-level BPE at its largest vocabulary size of $135k$ lies on the Pareto front, as it has the highest compression rate. On the other hand, both BLT and MYTE are Pareto-dominated. 

\subsection{Extrinsic Evaluation}

\subsubsection{English Classification Benchmarks}

{\setlength{\tabcolsep}{5.4pt}
\begin{table}[ht!]
    \centering
    \begin{tabular}{lccc}
        \toprule
        Model & PIQA & HellaSwag & Arc-C \\
        (size) & (50.00) & (25.00) & (25.00) \\
        \midrule
        BLT (4.5)        & 66.10 & 44.81 & 24.74 \\
        MYTE (90k)           & 67.14 & 44.47 & \underline{27.39} \\
        PA BPE (90k) & \underline{71.55} & \underline{53.22} & 26.19 \\
        BPE (90k)  & \textbf{72.31} & \textbf{54.42} & \textbf{28.41} \\
        \bottomrule
    \end{tabular}
    \caption{Accuracies on English classification benchmarks. Parenthesized values indicate the expected accuracy of a random classifier. The highest score is in \textbf{bold} and the second-highest is \underline{underlined}.}
    \label{tab:english-classification}
\end{table}
}

Table~\ref{tab:english-classification} shows that Byte-level BPE achieves the highest scores on all three English classification benchmarks. Its advantage is statistically significant across all comparisons, except over Parity-aware BPE on PIQA and MYTE on Arc-C, where its numerical lead does not reach significance. Parity-aware BPE is a strong runner-up, performing significantly better than BLT and MYTE on both PIQA and HellaSwag. For commonsense reasoning and sentence completion benchmarks, BPE-based tokenizers have a consistent advantage over alternative approaches in our evaluation setting. 

\subsubsection{Multilingual Classification Benchmarks}

{\setlength{\tabcolsep}{3.9pt}
\begin{table}[ht!]
    \centering
    \begin{tabular}{lccccc}
        \toprule
        Model & XNLI & XCOPA & XStoryCloze \\
        (size) & (33.33) & (50.00) & (50.00) \\
        \midrule
        BLT (4.5)              & 36.29 & 53.30 & 56.52 \\
        MYTE (90k)          & \textbf{42.49} & 54.93 & 50.55 \\
        PA BPE (90k) & 40.43 & \underline{58.70} & \underline{56.70} \\
        BPE (90k) & \underline{40.56} & \textbf{61.03} & \textbf{57.18} \\
        \bottomrule
    \end{tabular}
    \caption{Averaged per-language accuracies across multilingual classification benchmarks. Parenthesized values indicate the expected accuracy of a random classifier. The highest score is in \textbf{bold} and the second-highest is \underline{underlined}. Detailed results for each language are reported in Appendix~\ref{app:mcb}.}
    \label{tab:multilingual-classification}
\end{table}
}

Table~\ref{tab:multilingual-classification} reveals task-dependent performance across tokenizers, with no consistent winner across all three benchmarks. MYTE significantly outperforms all other tokenizers on XNLI, consistent with its morphological representations providing richer cross-lingual semantic signals for inference. Byte-level BPE achieves the highest scores on XCOPA and XStoryCloze, suggesting that its efficiency is best leveraged on tasks requiring causal and narrative reasoning.

\subsubsection{Machine Translation}

{\setlength{\tabcolsep}{6pt}
\begin{table}[ht!]
\centering
\begin{tabular}{lcc}
\toprule
Model (size) & EN $\to$ XX & XX $\to$ EN \\
\midrule
BLT (4.5)    & 10.82             & 11.70 \\
MYTE (90k)   & \textbf{14.77}    & \textbf{13.81} \\
PA BPE (90k) & 11.36             & 12.19 \\
BPE (90k)    & \underline{13.39} & \underline{12.78} \\
\bottomrule
\end{tabular}
\caption{Averaged BLEU scores across ten SEA languages. The highest score is in \textbf{bold} and the second-highest is \underline{underlined}.}
\label{tab:mt-bleu}
\end{table}
}

Table~\ref{tab:mt-bleu} aggregates the BLEU translation scores across ten SEA languages. We also provide the chrF scores in Appendix~\ref{app:mtc}. MYTE achieves the highest BLEU scores in both translation directions. A consistent directional asymmetry is observed across both metrics for MYTE, where it is systematically stronger in EN $\rightarrow$ XX translation than XX $\rightarrow$ EN. MYTE's morpheme-level segmentation enables finer-grained generation of morphologically complex target word forms in SEA languages, an advantage that narrows when translating into English. 

\section{Analysis}
\subsection{Effect of Scaling Vocabulary Size}
Increasing vocabulary size produces distinct behavior across tokenizers, as seen in Table~\ref{tab:tokenizer-comparison}. Byte-level BPE improves consistently across both efficiency and cross-lingual fairness with a larger vocabulary size. In contrast, BLT gains in efficiency but sacrifices cross-lingual equity as vocabulary size grows.

MYTE is relatively insensitive to morpheme inventory size scaling within the evaluated range. Compression rate and tokenizer parity remain nearly constant across all three morpheme inventory sizes, indicating that its morphological segmentation approach saturates at around 4,096 morphemes per language. 

Parity-aware BPE becomes more inequitable as vocabulary size increases. While it achieves the lowest Gini coefficient among all models, both its Gini coefficient and tokenizer parity worsen at larger vocabulary sizes, increasing to 0.029 and 1.25 respectively at $135k$ vocabulary size. 

\subsection{Fairness Regression in Parity-aware BPE}
It seems counterintuitive that Parity-aware BPE tokenizers are less equitable as vocabulary size increases. To investigate this, we analyzed per-language token counts produced by Parity-aware BPE tokenizers of different vocabulary sizes. We use the FLORES+ training dataset (rather than the test dataset) because it serves as the development corpus during tokenizer training. Examining this dataset would isolate the effect of vocabulary scaling and avoid confounding factors from unseen data such as differing vocabulary distributions.

\begin{figure}[ht!]
    \centering
    \includegraphics[width=1\columnwidth]{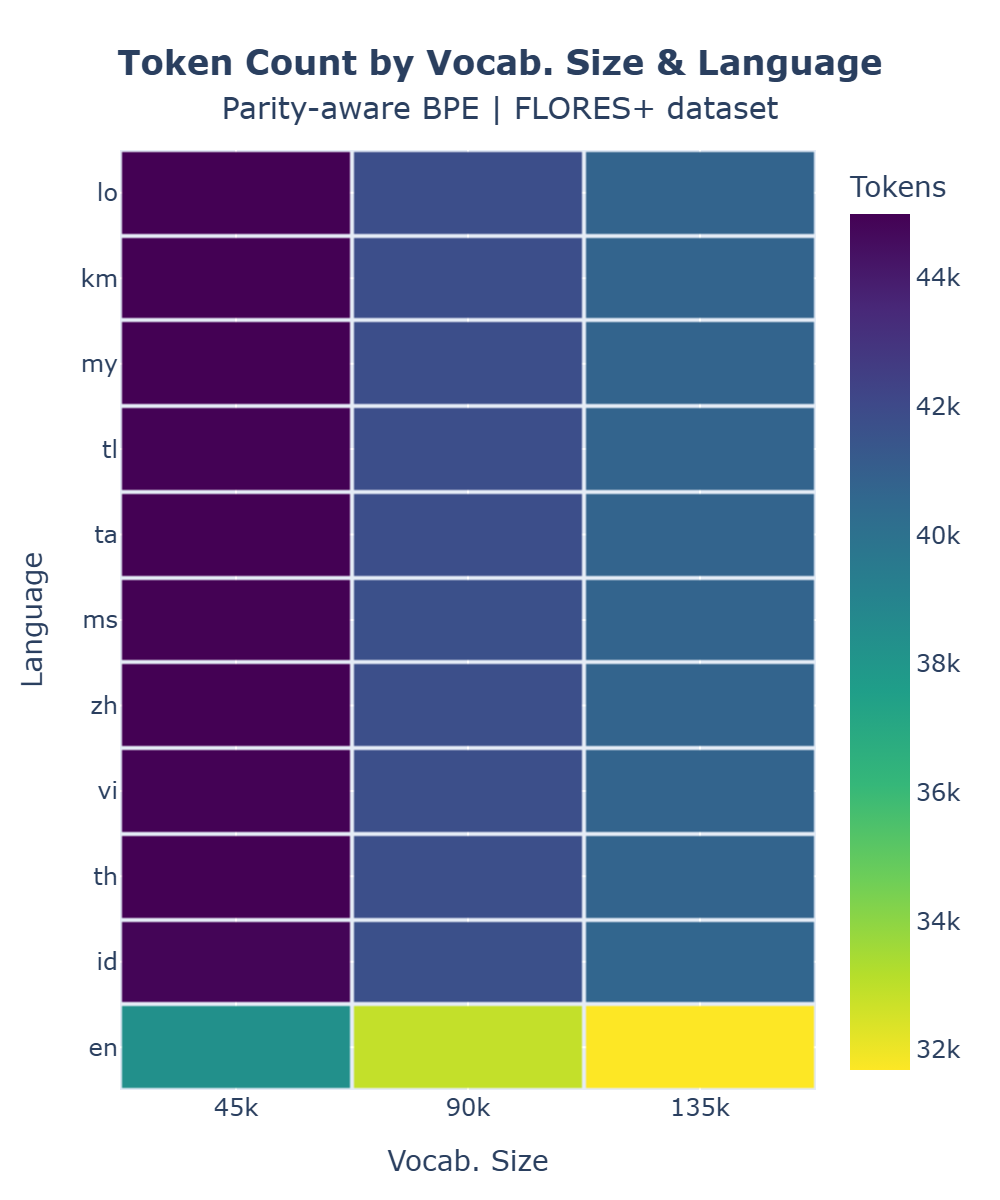}
    \caption{Per-language token counts on the FLORES+ training dataset by Parity-aware BPE tokenizers of varying vocabulary sizes.}
    \label{fig2}
\end{figure}

Figure~\ref{fig2} shows that increasing the vocabulary size results in a larger reduction in the tokens required for English sentences compared to the other SEA languages. This results in token count disparity between English and the other SEA languages to widen by 36\% as vocabulary size scales from $45k$ to $135k$, causing tokenizer parity to increase (i.e., worsen). We observe that Parity-aware BPE only limits worst-case per-language tokenizer parity, so its fairness mechanism does not prevent English from acquiring more merges as vocabulary size increases.

\subsection{Effect of Tokenizer Choice and Script Type}

To investigate whether tokenizer parity outcomes are driven by the choice of tokenizer, the script type of the target language, or their interaction, we apply a two-way mixed ANOVA test \citep{Meyers_Gamst_Guarino_2009}. Running separate pairwise $t$-tests for each tokenizer--script combination would inflate the Type I error rate multiplicatively.

The between-subject factor is script type, categorized as Latin (Indonesian, Malay, Tagalog, Vietnamese) or Abugida (Burmese, Khmer, Lao, Tamil, Thai) according to \citet{Limisiewicz_2024}. Chinese is excluded from this comparison as it is the only CJK-script language. The within-subject factor is tokenizer choice, as measuring the same language across tokenizers introduces within-subject correlation. We assess the effect of these factors on per-language tokenizer parity, measure statistical significance at $\alpha = 0.05$, and report partial $\eta^2$ as the effect size measure.

\begin{table}[ht!]
    \centering
    \begin{tabular}{lcc}
        \toprule
        Tokenizer & \multicolumn{2}{c}{Script type} \\ \cline{2-3}
        (size) & Latin & Abugida \\
        \midrule
        BLT (4.5)             & 2.36 & 3.38 \\
        MYTE (90k)             & \textbf{1.19} & \underline{1.32} \\
        PA BPE (90k) & \underline{1.26} & \textbf{1.22} \\
        BPE (90k)   & 1.30 & 2.18 \\
        \bottomrule
    \end{tabular}
    \caption{Per-language tokenizer parity, macro-averaged by tokenizer and script type. Latin scripts exclude English. The lowest value is in \textbf{bold} and the second-lowest is \underline{underlined}.}
    \label{tab:parity-script}
\end{table}

Tokenizer choice is the only statistically significant factor affecting tokenizer parity ($p < 0.001$, partial $\eta^2 = 0.752$). The large effect size indicates that tokenizer choice explains the majority of variance in tokenizer parity, regardless of script type. At the same time, script type alone does not reach significance ($p = 0.090$). As shown in Table~\ref{tab:parity-script}, Parity-aware BPE and MYTE achieve near-uniform tokenizer parity across both script types. In contrast, Byte-level BPE imposes a 1.68× tokenizer parity penalty on Abugida scripts relative to Latin scripts. BLT also exhibits a high cross-script disparity (1.43×), consistent with its entropy model being undertrained on non-Latin scripts. 

Fundamentally, tokenizer parity directly determines inference cost. A tokenizer parity of $k$ for a given language implies that a user pays $k$ times the per-token API cost relative to English to process the same semantic content. Under Byte-level BPE, Abugida-script users incur 1.68× higher costs on average than English users for semantically equivalent prompts. These results confirm that equitable tokenizer design, and not script similarity to English, is the main determinant of whether a tokenizer imposes uniform computational costs across languages.

\section{Conclusion}
We present the first systematic, dataset-controlled comparison of BLT, MYTE, Parity-aware BPE, and Byte-level BPE across eleven SEA languages, evaluating tokenizer equity, compression efficiency, and downstream task performance under the same experimental conditions. Our intrinsic evaluation demonstrates that cross-lingual equity and tokenization efficiency are not fundamentally at odds. 

Among the equitable tokenizers analyzed, MYTE delivers the strongest semantic inference and machine translation performance through richer morphological representations, though at the cost of a higher computational budget and lower compression efficiency. Despite its architectural novelty in eliminating fixed tokenizer vocabulary, we found that BLT underperforms on downstream tasks as its entropy model receives insufficient exposure to low-resource languages under realistic multilingual data distribution.

The appropriate choice of tokenizer is ultimately use-case dependent. We recommend Parity-aware BPE as a responsible default for multilingual models targeting SEA languages, given its favorable position in the efficiency-equity space and relatively strong downstream task performance. However, the base variant of Parity-aware BPE requires multi-way parallel data, which may be scarce or unavailable in low-resource settings \citep{foroutan2025parityaware}. MYTE is preferred when morphology and translation are critical, and the computational budget permits the associated training overhead. 

Our ANOVA results establish that tokenizer choice is the primary factor affecting tokenizer parity. The difference in inference costs between English and SEA-language users can be addressed through the choice of the tokenizer. Equitable tokenization has direct, quantifiable consequences for the 671 million speakers across Southeast Asia \citep{lovenia2025seacrowdmultilingualmultimodaldata}, for whom token-priced APIs create unequal access costs. How tokenization is carried out across languages shapes the economic accessibility of multilingual models for underrepresented language communities.

\section*{Limitations}
First, all language models are trained at the 1.5B-parameter scale due to computational resource constraints. We leave the investigation of larger model sizes to future work. Next, BLT cannot be scaled along the same vocabulary dimension as the other tokenizers since it operates without a fixed vocabulary, which prevents direct matched-vocabulary size comparison. Finally, we evaluate only base pretrained models. We believe this is sufficient, as supervised fine-tuning or alignment training does not affect the tokenizer’s fairness or efficiency.

We do not anticipate any immediate societal or individual harm arising from this work. Nevertheless, we advise users to exercise caution, as our models have not been subjected to safety or value alignment procedures.

\bibliography{custom}

\clearpage
\appendix

\section{Training Data Details}
\subsection{Tokenizer Training Data} \label{app:tok-training-data}
\begin{table}[ht!]
    \centering
    \begin{tabular}{llr}
        \toprule
        ISO & Language & \# Sentences \\
        Code & & \\
        \midrule
        en & English    & 604,793 \\
        id & Indonesian & 129,906 \\
        th & Thai       & 115,159 \\
        vi & Vietnamese &  90,470 \\
        zh & Chinese    &  25,683 \\
        ms & Malay      &  21,872 \\
        ta & Tamil      &   5,799 \\
        tl & Tagalog    &   3,480 \\
        my & Burmese    &   1,349 \\
        km & Khmer      &   1,253 \\
        lo & Lao        &     236 \\
        \midrule
        \textbf{Total} & & \textbf{1,000,000} \\
        \bottomrule
    \end{tabular}%
    \caption{Per-language sentence counts in the tokenizer training dataset.}
    \label{tab:tok-training-data}
\end{table}

\subsection{Language Model Training Data}
\label{app:lm-training-data}

We source pretraining data from the FineWeb2 corpus \citep{penedo2025fineweb2pipelinescale}, randomly sampling a subset of 100 million sentences (203 GB) spanning all eleven SEA languages. To balance coverage between high-resource and low-resource languages, we control the language proportions via temperature sampling, following the approach of \citet{foroutan2025parityaware}.

\paragraph{Temperature Sampling} Sampling each language proportionally to its word count in FineWeb2 would overwhelmingly favor English at 94.3\%. As such, we sample according to a temperature-scaled probability:
\begin{equation}
    p(L) \propto |L|^{1/\tau},
    \label{eq:temp-sampling}
\end{equation}
where $p(L)$ is the probability of sampling text from language $L$ during pre-training, $|L|$ is the number of words in that language in the corpus, and $\tau$ is a temperature parameter. When $\tau = 1$, sampling is purely proportional to word frequency. As $\tau$ increases, the distribution becomes increasingly uniform, thereby boosting the relative sampling probability of low-resource languages.

We configure $\tau = 1.21$ so that English constitutes 89.7\% of the resulting training sentences, matching the proportion of English data used in the Llama 2 pretraining dataset \citep{touvron2023llama2openfoundation}. Table~\ref{tab:lm-training-data-raw} shows the raw word frequency of each language in FineWeb2, along with the adjusted frequency after temperature sampling is applied. For each language $L$, we compute $|L|^{1/\tau}$ and normalize across all eleven SEA languages to obtain the final sampling proportion. These proportions are then used to determine the number of sentences drawn from each language in our 100M-sentence subset (Table~\ref{tab:lm-training-data}).

\begin{table*}[ht!]
    \centering
    \begin{tabular}{lrrr}
        \toprule
        Language
        & \makecell{Word frequency (billion), $|L|$}
        & \makecell{Relative frequency, $|L|^{1/\tau}$}
        & Proportion \\
        \midrule
        English    & 11,500.0 & 2269.6 & 89.68\% \\
        Chinese    &    543.5 &  182.2 &  7.20\% \\
        Indonesian &     60.3 &   29.6 &  1.17\% \\
        Vietnamese &     50.9 &   25.7 &  1.02\% \\
        Thai       &     24.7 &   14.1 &  0.56\% \\
        Malay      &      5.6 &   4 .2 &  0.17\% \\
        Tamil      &      1.9 &    1.7 &  0.07\% \\
        Tagalog    &      1.6 &    1.5 &  0.06\% \\
        Burmese    &      0.9 &    0.9 &  0.03\% \\
        Khmer      &      0.7 &    0.7 &  0.03\% \\
        Lao        &      0.2 &    0.3 &  0.01\% \\
        \midrule
        \textbf{Total} & \textbf{12,190.3} 
            & \textbf{2,530.5} & \textbf{100.00\%} \\
        \bottomrule
    \end{tabular}
    \caption{Word frequencies in FineWeb2 and relative frequencies after temperature sampling ($\tau=1.21$). Temperature sampling boosts the relative proportion of low-resource SEA languages while keeping English at 89.7\%, matching Llama~2's pretraining data distribution.}
    \label{tab:lm-training-data-raw}
\end{table*}

\begin{table}[ht!]
    \centering
    \begin{tabular}{llr}
        \toprule
        ISO & Language & \# Sentences \\
        Code & & \\
        \midrule
        en & English    & 89,689,566 \\
        zh & Chinese    &  7,199,747 \\
        id & Indonesian &  1,169,277 \\
        vi & Vietnamese &  1,016,743 \\
        th & Thai       &    558,780 \\
        ms & Malay      &    165,279 \\
        ta & Tamil      &     68,252 \\
        tl & Tagalog    &     59,364 \\
        my & Burmese    &     34,699 \\
        km & Khmer      &     28,295 \\
        lo & Lao        &      9,998 \\
        \midrule
        \textbf{Total} & & \textbf{100,000,000} \\
        \bottomrule
    \end{tabular}%
    \caption{Per-language sentence counts in the language model training dataset after temperature sampling.}
    \label{tab:lm-training-data}
\end{table}

\subsection{Machine Translation Fine-tuning} \label{app:ft-training-data}

Machine translation fine-tuning is performed via continual pre-training on English-XX parallel sentences for one epoch. The fine-tuning dataset consists of up to 13 million parallel sentence pairs per language, which were randomly sampled without replacement from NLLB \citep{costajussa2024nllb} where possible, following the approach of \citet{nguyen2026opensealgoodfastcheap}. The resulting per-language sentence counts are shown in Table~\ref{tab:mttdstats}.

\begin{table}[ht!]
    \centering
    \begin{tabular}{llr}
        \toprule
        ISO & Language & \# Sentences \\
        Code & & (million) \\
        \midrule
        zh & Chinese    & 13.0 \\
        id & Indonesian & 13.0 \\
        vi & Vietnamese & 13.0 \\
        th & Thai       & 13.0 \\
        ms & Malay      & 13.0 \\
        ta & Tamil      & 13.0 \\
        tl & Tagalog    & 13.0 \\
        my & Burmese    & 10.0 \\
        km & Khmer      &  5.8 \\
        lo & Lao        &  4.2 \\
        \midrule
        \textbf{Total} & & \textbf{111.0} \\
        \bottomrule
    \end{tabular}%
    \caption{Per-language sentence counts in the fine-tuning dataset.}
    \label{tab:mttdstats}
\end{table}

Each parallel sentence pair is formatted using the template by \citet{qorib2025justparallelimprovingmultilingual}: \texttt{"\{source language\}: \{source sentence\} \textbackslash n \{target language\}: \{target} \texttt{sentence\}"}. To prevent the model from developing a bias toward a fixed source language, the language order of each pair was randomized independently with equal probability. This means each example has an equal chance of being presented as English-first or target language-first. 

\section{Intrinsic Tokenizer Evaluation Metrics} \label{intrin-tok}

\subsection{Cross-lingual Equity Metrics}

\textbf{Tokenizer parity} measures the ratio of the average number of tokens per sentence in a given language relative to English \citep{petrov2023languagemodeltokenizersintroduce}. For a specific language $L$ with $k$ aligned sentences, we can compute its average number of tokens per sentence, $average\_tokens_{L}$:
\begin{equation}
\frac{1}{k} \sum_{i=1}^{k} \text{\# tokens~in~sentence}~i
\end{equation}

The tokenizer parity for language $L$, $parity_{L}$, is defined as:
\begin{equation}
\frac{average\_tokens_{L}}{average\_tokens_{English}}
\end{equation}

The macro-average tokenizer parity for all $n$ non-English languages is computed as:
\begin{equation}
\frac{1}{n} \sum_{j=1}^{n}parity_{j} 
\end{equation}
This ratio measures whether tokenizers impose computational costs unequally across languages. A macro-average tokenizer parity \emph{closer to 1} indicates a more equitable tokenizer across languages.

\textbf{Gini coefficient} assesses tokenization equity by treating token costs as a distribution \citep{foroutan2025parityaware}. The token cost for a language, $c$, is defined as the average number of tokens per sentence for the language in the parallel corpus. For token costs \( c_1 \leq c_2 \leq \dots \leq c_n \) across $n$ languages, the Gini coefficient is computed as:
\begin{equation}
\frac{1}{n}\!\left(n + 1 - 2\,\frac{ \sum_{i=1}^{n} (n + 1 - i)\,c_i}{\sum_{i=1}^{n} c_i}\right)
\end{equation}
Values range from 0 to 1. A \emph{lower Gini coefficient} (closer to 0) indicates a more equitable tokenizer across languages. 

\subsection{Tokenizer Efficiency Metrics}

\textbf{Compression rate} measures how efficiently a tokenizer compresses text. It is defined as the average of the inverse token count per sentence \citep{foroutan2025parityaware}. For a specific language $L$, its compression rate, $rate_{L}$ is computed as:
\begin{equation}
\frac{1}{k} \sum_{i=1}^{k} \frac{1}{\text{\# tokens~in~sentence}~i}
\end{equation}
where $k$ is the number of aligned sentences for language $L$ in the parallel corpus. Essentially, we compute a language's compression rate by evaluating the inverse of the number of tokens per sentence and then averaging it.

The macro-average compression rate for all $n$ languages is computed as:
\begin{equation}
\frac{1}{n} \sum_{j=1}^{n}rate_{j} 
\end{equation}

Utilizing a parallel corpus controls for semantic differences, by comparing token counts over semantically equivalent content. A \emph{higher macro-average compression rate} indicates a more efficient tokenizer across languages. This metric is informative when viewed alongside tokenizer parity, as a high overall compression rate can mask under-compression of individual low-resource languages.

\section{Extrinsic Metrics} \label{ext-met}
\subsection{English Classification Benchmarks}

Classification benchmarks evaluate a model's ability to understand, analyze, and select the correct category from a set of options. The following English classification benchmarks are used by \citet{pagnoni2024bytelatenttransformerpatches} to evaluate a model's commonsense reasoning and general world knowledge.

\textbf{Physical Intuition Question Answering (PIQA)} \citep{bisk2019piqareasoningphysicalcommonsense} probes a model's understanding of everyday physical interactions and how objects behave in the real world. Each example presents a goal and two solution candidates, with the model tasked to identify the more physically plausible option.

\textbf{HellaSwag} \citep{zellers2019hellaswagmachinereallyfinish} is a commonsense natural language inference benchmark where a model must select the most plausible continuation of a given scenario from four candidate endings. The dataset is constructed using adversarial filtering to ensure that a model possesses genuine contextual understanding.

\textbf{Arc-Challenge (Arc-C)} \citep{clark2018thinksolvedquestionanswering} evaluates a model's scientific reasoning ability through multiple-choice questions drawn from grade-school science exams. The Challenge subset selects questions that simple retrieval-based and word co-occurrence methods fail to answer correctly, making it a reliable indicator of deeper reasoning capabilities.

\subsection{Multilingual Classification Benchmarks} \label{mcb}

The following multilingual classification benchmarks are used by \citet{foroutan2025parityaware} and they collectively span several of our target languages. They enable a comprehensive evaluation of a model's cross-lingual performance on SEA languages.

\textbf{Cross-lingual Natural Language Inference (XNLI)} \citep{conneau2018xnlievaluatingcrosslingualsentence} extends the MultiNLI dataset to 15 languages and serves as a standard benchmark for cross-lingual natural language understanding. Models must classify the logical relationship between each pair as one of three categories: entailment, contradiction, or neutral. This benchmark covers English, Chinese, Thai, and Vietnamese.

\textbf{Cross-lingual Choice of Plausible Alternatives (XCOPA)} \citep{ponti2020xcopamultilingualdatasetcausal} is a multilingual benchmark targeting causal commonsense reasoning. Given a premise, a model must identify either the most plausible cause or effect from two candidate sentences. XCOPA is evaluated in a zero-shot setting to assess cross-lingual transfer without fine-tuning. This benchmark covers English, Chinese, Indonesian, Tamil, Thai, and Vietnamese.

\textbf{XStoryCloze} \citep{lin2022fewshotlearningmultilinguallanguage} requires a model to select the correct ending for a four-sentence narrative from two candidate conclusions in a multilingual setting. It evaluates cross-lingual narrative understanding and commonsense reasoning. This benchmark covers English, Burmese, Chinese, and Indonesian.

\subsection{Machine Translation} \label{mt}

Machine translation is a natural benchmark for evaluating multilingual LLMs, as it tests a model's ability to understand and generate text across languages. For SEA languages, translation quality serves as a proxy for how well a model has internalized low-resource linguistic structure \citep{issaka2026translationscalableproxymultilingual}.

Machine translation performance is measured by comparing the machine-generated output to human reference translations. Two complementary metrics are typically used, BLEU (Bilingual Evaluation Understudy) \citep{papineni-etal-2002-bleu} and chrF (character-level F-score) \citep{popovic-2015-chrf}. Both metrics have scores ranging from 0 to 100, with \emph{higher scores} indicating better translation quality. BLEU measures word-level n-gram precision against reference translations and was used by \citet{pagnoni2024bytelatenttransformerpatches}. chrF computes character n-gram F-score and is suitable for morphologically rich languages where word-level overlap may be sparse and was used by \citet{Limisiewicz_2024}. 

The detailed BLEU and chrF translation scores are shown in Appendix~\ref{app:mt0}. The original MYTE paper \citep{Limisiewicz_2024} reports scores only for English-to-Vietnamese and English-to-Tamil translation among SEA languages, and our scores are higher than their reported scores.

\section{Detailed Results}
\subsection{Multilingual Classification Benchmarks} \label{app:mcb}

\begin{table}[ht]
    \centering
    \makebox[\textwidth][c]{%
        \begin{minipage}{\textwidth}
            \centering
                \begin{tabular}{lccccc}
                \toprule
                Model & en & zh & vi & th & AVG \\
                (size) & & & & & \\
                \midrule
                BLT (4.5)         & 42.10 & 33.99 & 34.29 & 34.79 & 36.29 \\
                MYTE (90k)           & 50.00 & 33.99 & 44.99 & 40.98 & 42.49 \\
                PA BPE (90k) & 47.94 & 33.91 & 43.07 & 36.81 & 40.43 \\
                BPE (90k)   & 49.30 & 33.51 & 43.09 & 36.35 & 40.56 \\
                \bottomrule
            \end{tabular}%
            \caption{Per-language XNLI scores. Expected accuracy of a random classifier = 33.33.}
            \label{tab:xnli}
        \end{minipage}%
    }
\end{table}

\begin{table}[ht]
    \makebox[\textwidth][c]{%
        \begin{minipage}{\textwidth}
            \centering
            \begin{tabular}{lccccccc}
                \toprule
                Model & en & zh & id & vi & th & ta & AVG \\
                (size) & & & & & & & \\
                \midrule
                BLT (4.5)              & 64.20 & 52.40 & 52.60 & 48.60 & 52.60 & 49.40 & 53.30 \\
                MYTE (90k)             & 54.40 & 51.00 & 55.80 & 56.60 & 55.00 & 56.80 & 54.93 \\
                PA BPE (90k) & 71.40 & 55.60 & 58.60 & 60.20 & 53.40 & 53.00 & 58.70 \\
                BPE (90k)   & 71.60 & 59.00 & 61.60 & 62.80 & 56.20 & 55.00 & 61.03 \\
                \bottomrule
            \end{tabular}
            \caption{Per-language XCOPA scores. Expected accuracy of a random classifier = 50.00.}
            \label{tab:xcopa}
        \end{minipage}%
    }
\end{table}

\begin{table}[ht]
    \makebox[\textwidth][c]{%
        \begin{minipage}{\textwidth}
            \centering
            \begin{tabular}{lccccc}
                \toprule
                Model & en & zh & id & my & AVG \\
                (size) & & & & & \\
                \midrule
                BLT (4.5)              & 65.45 & 54.20 & 55.06 & 51.36 & 56.52 \\
                MYTE (90k)             & 52.95 & 49.83 & 50.89 & 48.51 & 50.55 \\
                PA BPE (90k) & 65.25 & 55.06 & 55.92 & 50.56 & 56.70 \\
                BPE (90k)   & 65.78 & 54.80 & 57.64 & 50.50 & 57.18 \\
                \bottomrule
            \end{tabular}
            \caption{Per-language XStoryCloze scores. Expected accuracy of a random classifier = 50.00.}
            \label{tab:xsc}
        \end{minipage}%
    }
\end{table}

\clearpage
\subsection{Machine Translation} \label{app:mt0}
\subsubsection{BLEU scores} \label{app:mtb}

{\setlength{\tabcolsep}{5pt}
\begin{table}[ht!]
    \makebox[\textwidth][c]{%
        \begin{minipage}{\textwidth}
            \centering
            \begin{tabular}{lcccccccccccc}
                \toprule
                Model & zh & id & vi & th & ms & ta & tl & my & km & lo & AVG \\
                (size) & & & & & & & & & & & & \\
                \midrule
                BLT (4.5)              & 9.18  & 27.01 & 19.30 & 7.24 & 24.98 & 3.65 & 11.04 & 1.92 & 2.31 & 1.53 & 10.82 \\
                MYTE (90k)             & 18.43 & 34.92 & 25.36 & 9.25 & 27.66 & 5.61 & 16.73 & 2.49 & 3.40 & 3.90 & 14.77 \\
                PA BPE (90k) & 22.78 & 26.47 & 18.30 & 5.64 & 21.04 & 4.38 & 9.11  & 1.30 & 2.56 & 2.01 & 11.36 \\
                BPE (90k)   & 30.20 & 27.72 & 32.51 & 6.90 & 23.10 & 2.56 & 8.05  & 0.58 & 1.45 & 0.82 & 13.39 \\
                \bottomrule
            \end{tabular}%
            \caption{Per-language BLEU scores (EN $\rightarrow$ XX).}
            \label{tab:bleu_en_xx}
        \end{minipage}%
    }
\end{table}}

{\setlength{\tabcolsep}{5pt}
\begin{table}[ht!]
    \makebox[\textwidth][c]{%
        \begin{minipage}{\textwidth}
            \centering
            \begin{tabular}{lccccccccccccc}
                \toprule
                Model & zh & id & vi & th & ms & ta & tl & my & km & lo & AVG \\
                (size) & & & & & & & & & & & & \\
                \midrule
                BLT (4.5)              & 10.33 & 25.63 & 21.94 & 8.47  & 18.30 & 4.65 & 17.10 & 3.92 & 4.31 & 2.33 & 11.70 \\
                MYTE (90k)             & 16.65 & 30.96 & 18.71 & 13.70 & 23.90 & 3.76 & 21.23 & 3.87 & 2.40 & 2.90 & 13.81 \\
                PA BPE (90k) & 14.75 & 26.06 & 22.82 & 12.17 & 17.29 & 5.35 & 16.73 & 2.15 & 2.10 & 2.47 & 12.19 \\
                BPE (90k)   & 14.42 & 29.75 & 23.24 & 11.26 & 19.20 & 5.62 & 15.05 & 2.39 & 3.13 & 3.77 & 12.78 \\
                \bottomrule
            \end{tabular}%
            \caption{Per-language BLEU scores (XX $\rightarrow$ EN).}
            \label{tab:bleu_xx_en}
        \end{minipage}%
    }
\end{table}}

\subsubsection{chrF scores} \label{app:mtc}

{\setlength{\tabcolsep}{4pt}
\begin{table}[ht!]
    \makebox[\textwidth][c]{%
        \begin{minipage}{\textwidth}
            \centering
            \begin{tabular}{lcccccccccccc}
                \toprule
                Model & zh & id & vi & th & ms & ta & tl & my & km & lo & AVG \\
                (size) & & & & & & & & & & & & \\
                \midrule
                BLT (4.5)             & 13.16 & 48.75 & 42.03 & 30.48 & 45.46 & 31.52 & 31.35 & 19.91 & 19.92 & 20.63 & 30.32 \\
                MYTE (90k)             & 44.13 & 59.82 & 54.22 & 40.37 & 47.27 & 26.22 & 42.02 & 22.46 & 26.29 & 25.89 & 38.87 \\
                PA BPE (90k) & 15.29 & 57.87 & 46.46 & 24.16 & 51.92 & 32.62 & 44.76 & 20.00 & 17.82 & 19.58 & 33.05 \\
                BPE (90k)   & 27.86 & 67.47 & 54.80 & 30.81 & 62.05 & 30.07 & 51.56 & 14.89 & 15.87 & 13.80 & 36.92 \\
                \bottomrule
            \end{tabular}%
            \caption{Per-language chrF scores (EN $\rightarrow$ XX).}
            \label{tab:chrf_en_xx}
        \end{minipage}%
    }
\end{table}}

{\setlength{\tabcolsep}{4pt}
\begin{table}[ht!]
    \makebox[\textwidth][c]{%
        \begin{minipage}{\textwidth}
            \centering
            \begin{tabular}{lcccccccccccc}
                \toprule
                Model & zh & id & vi & th & ms & ta & tl & my & km & lo & AVG \\
                (size) & & & & & & & & & & & & \\
                \midrule
                BLT (4.5)              & 31.26 & 54.78 & 43.12 & 34.10 & 44.98 & 23.80 & 36.71 & 19.33 & 19.54 & 18.05 & 32.57 \\
                MYTE (90k)             & 28.18 & 60.04 & 40.16 & 31.13 & 59.96 & 23.45 & 42.58 & 19.38 & 20.56 & 20.32 & 34.58 \\
                PA BPE (90k) & 41.60 & 51.99 & 50.27 & 33.43 & 42.87 & 25.37 & 36.85 & 20.05 & 20.66 & 21.99 & 34.51 \\
                BPE (90k)   & 43.16 & 62.22 & 53.37 & 35.38 & 45.52 & 28.89 & 38.39 & 21.93 & 24.15 & 24.30 & 37.73 \\
                \bottomrule
            \end{tabular}%
            \caption{Per-language chrF scores (XX $\rightarrow$ EN).}
            \label{tab:chrf_xx_en}
        \end{minipage}%
    }
\end{table}}

\end{document}